\documentclass[a4paper,11pt,final]{article}

\usepackage{graphicx,amsmath,amssymb,url} 
\usepackage{epsfig}

\begin{document}

\title{ \LARGE\bf Wrong Colored Vermeer: Color-Symmetric Image Distortion}

\author{Hendrik Richter \\
HTWK Leipzig University of Applied Sciences \\ D--04251 Leipzig, Germany. \\ Email: 
hendrik.richter@htwk-leipzig.de. }

\maketitle

\begin{abstract}
Color symmetry implies that the colors of  geometrical objects are assigned according to their symmetry properties. 
It is defined by associating the elements of the symmetry group with a color permutation.
I use this concept for generative art and apply symmetry-consistent color distortions to images of paintings by Johannes Vermeer.  The color permutations are realized as mappings of the HSV color space onto itself.
\end{abstract}

\section*{Color Symmetry}
Symmetry is an ubiquitous concept in many branches of mathematics, most clearly visible, for instance, in geometry or group theory or mathematical physics. Symmetry is even more frequently present in visual arts, for example in architecture or painting or drawing. Thus, juxtaposing mathematics and art immediately suggest dealing with symmetry.  In the following, I focus on a special kind of symmetry: color symmetry.     

Suppose we have a geometrical object with some symmetry properties. This implies that the object in invariant under certain transformations. A complete set of these transformations yields the symmetry group $\mathcal{G}$ of the geometrical object. Further assume that the geometrical object is (or can be) colored. Thus, color symmetry  means that the coloring of the object is consistent with the symmetry group. More specifically, an element $g \in \mathcal{G}$ is called a color symmetry if $g$ is associated with a color permutation over a color set~\cite{roth82}.  In other words, $g$ permutes the colors  consistently by a color permutation assigning which elements of $\mathcal{G}$ change and which elements of $\mathcal{G}$ preserve colors,  see for instance~\cite{ada16,ada16a,farris17,rich20,rich21} for explicitly addressing the topic of color symmetry in generative art. In principle, a color symmetry may change all colors or even preserve all colors. From an artistic point of view, however, color symmetry is more interesting if we have a permutation which mixes color-changing and color-preserving elements. 

\begin{figure}[htb]
\centering
	\includegraphics[trim = 40mm 30mm 40mm 25mm,clip,width=4cm]{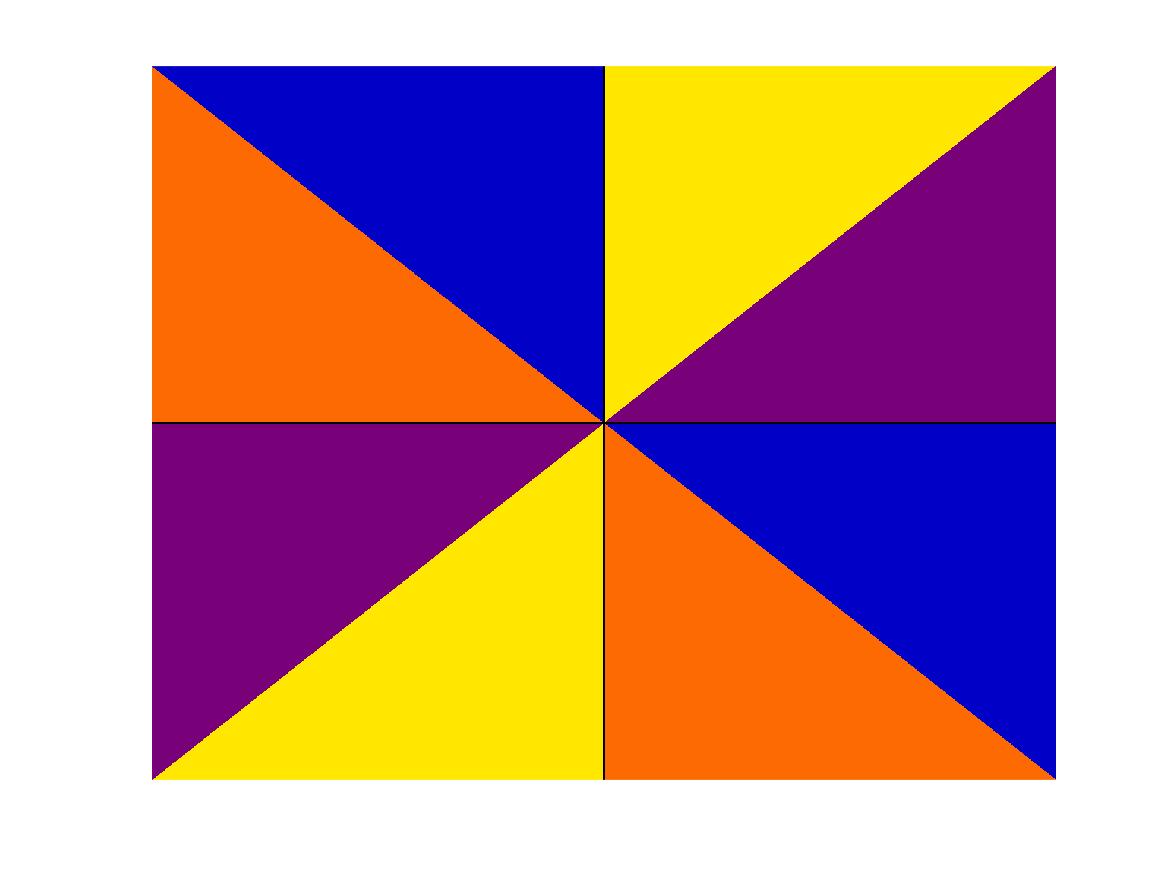}
    	\includegraphics[trim = 40mm 30mm 40mm 25mm,clip,width=4cm]{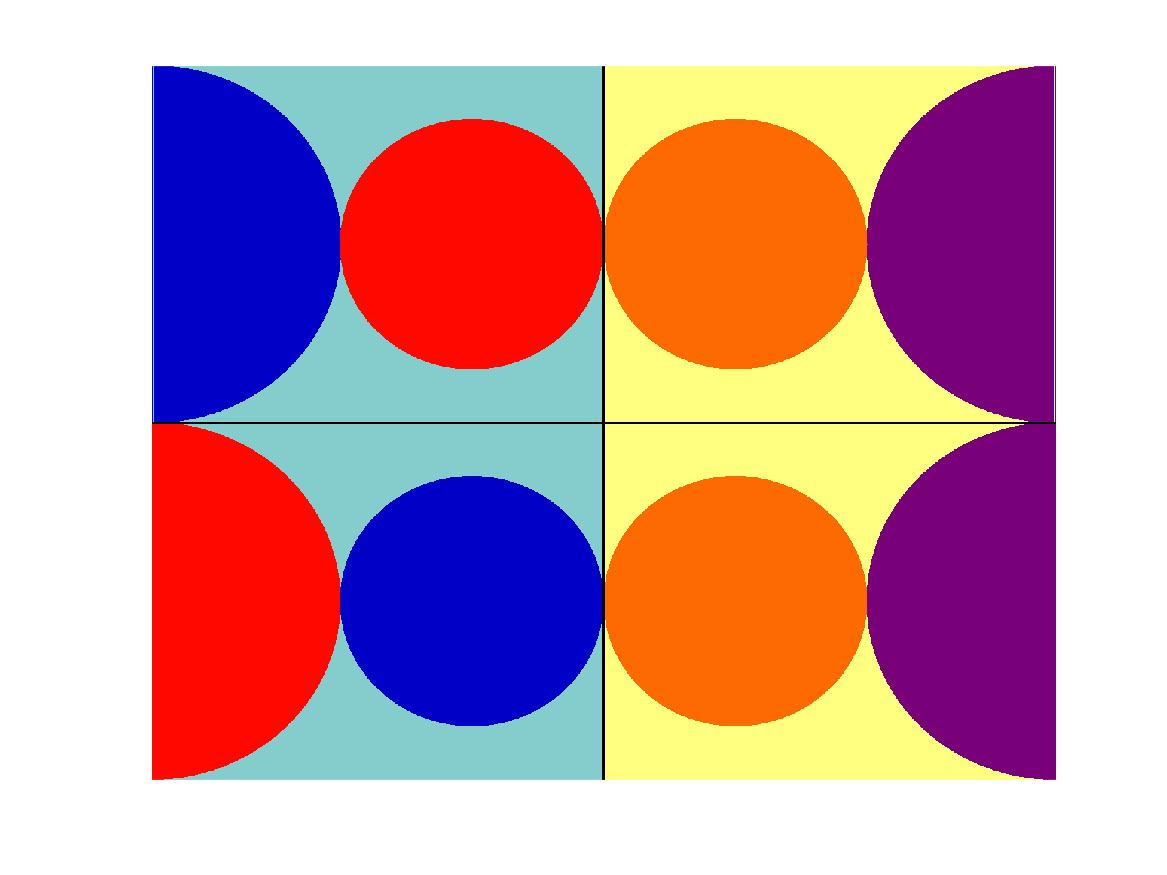}
	\includegraphics[trim = 40mm 30mm 40mm 25mm,clip,width=4cm]{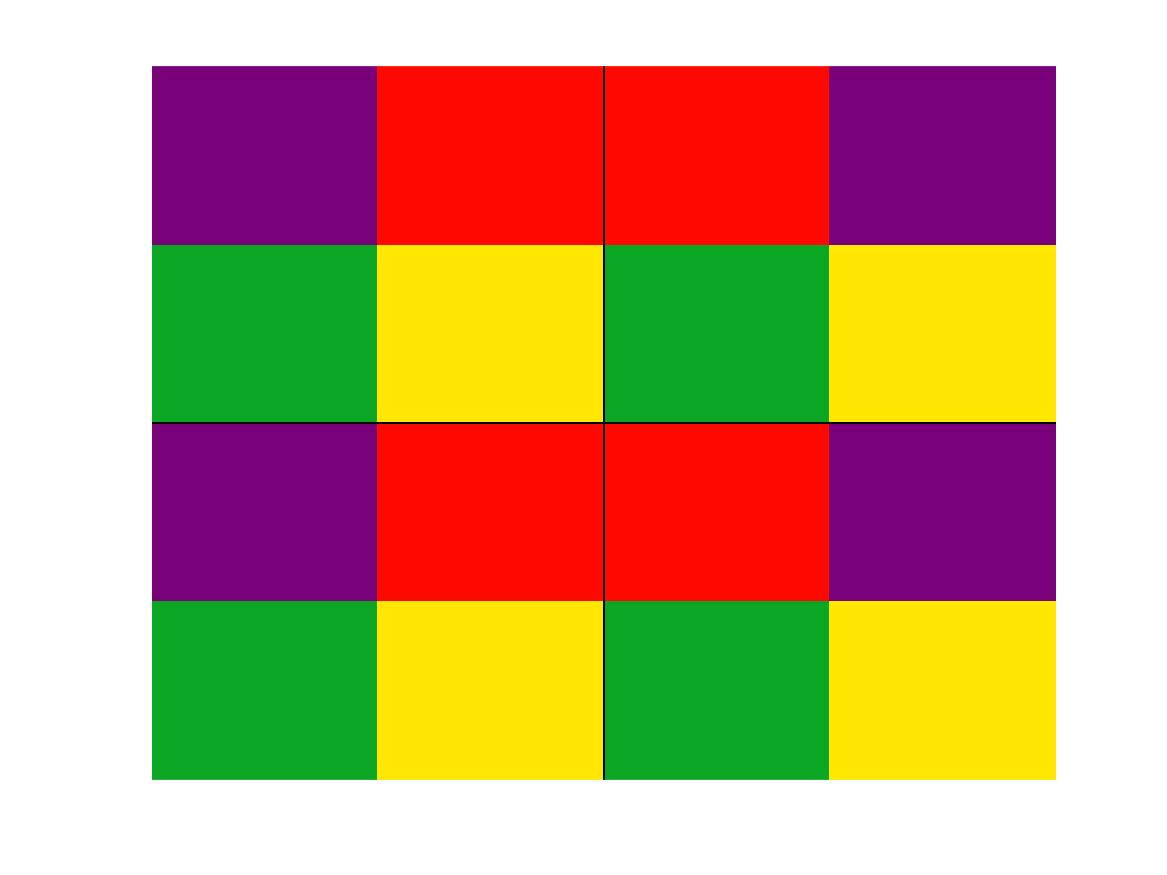}
\caption{Simple color symmetry.}
\label{fig:simplecolorsymmetry}
\end{figure}

Take as a very simple example a 2D rectangular object as in Figure~\ref{fig:simplecolorsymmetry}.~The symmetry group of the rectangle is the dihedral group $D_2$ of order $4$. A rotation $\alpha$ turns the rectangle counter-clockwise  180$^\circ$, the reflections $\beta_h$ and $\beta_v$ mirror the rectangle in the horizontal and vertical axis. The composed symmetries are from left to right, as for instance $\beta_h \beta_v= \alpha$ is a horizontal reflection followed by a vertical reflection, which equals a rotation. With the identity $e=\alpha^2=\beta_h^2=\beta_v^2$ and $\beta_h \beta_v=\beta_v \beta_h= \alpha$, we obtain the symmetry group 
$\mathcal{G}=(e,\alpha,\beta_h, \beta_v)$. The reflection axes divide the image into 4 sections, north-east (\textbf{ne}), north-west (\textbf{nw}), south-east (\textbf{se}) and south-west (\textbf{sw}). Each section can be mapped into any other section by the elements of the symmetry group. A rotation maps (\textbf{nw},\textbf{ne}) $\leftrightarrow$ (\textbf{se},\textbf{sw}), while for a horizontal and a vertical reflection the mapping is (\textbf{nw},\textbf{ne}) $\leftrightarrow$ (\textbf{sw},\textbf{se}) and (\textbf{nw},\textbf{sw}) $\leftrightarrow$ (\textbf{ne},\textbf{se}), respectively. 
 The sections in Figure~\ref{fig:simplecolorsymmetry} are further subdivided into subsections, which can be individually colored. Thus, the rectangles show simple color symmetries. For instance, in the left image in Figure~\ref{fig:simplecolorsymmetry} we have triangular subsections and a rotation $\alpha$ is associated with preserving the colors yellow $\mathbf{(y)}$ and purple $\mathbf{(p)}$ and changing blue $\mathbf{(b)}$ to orange $\mathbf{(o)}$ and orange to blue.  Using Cauchy's two-line notation, we may describe the color permutation $p_\alpha$ associated with the rotation $\alpha$ by
 $p_\alpha=\left(\begin{smallmatrix} \mathbf{o} & \mathbf{b} & \mathbf{y} & \mathbf{p}  \\    \mathbf{b} & \mathbf{o} & \mathbf{y} & \mathbf{p}\end{smallmatrix} \right)$.
  Both reflections change all colors: blue $\leftrightarrow$ yellow, orange $\leftrightarrow$ purple, purple $\leftrightarrow$ blue and yellow $\leftrightarrow$ orange. The color permutations $p_{\beta_h}$ and $p_{\beta_v}$ associated with the reflections (horizontal and vertical)  can be expressed similarly by a two-line notation. 
 
 Color symmetry may likewise apply to other schemes of subdividing the sections of a rectangle, for instance bubble-like or chessboard-like  subsections.
 The middle and the right image of Figure~\ref{fig:simplecolorsymmetry} shows further examples of color symmetry where the horizontal (middle image) and vertical (right image) reflection preserve some colors, while the rotation changes all colors.  The schemes of subdividing the sections may be different in the geometrical form (triangle, circle, rectangle, etc.) or in the granularity. The granularity is expressed by the number of subsection $\Lambda$. For instance, we have a granularity of $\Lambda=\{8,12,16\}$ with the 8, 12 and 16 subsections in the images in Figure~\ref{fig:simplecolorsymmetry} (from left to right).  In the following I propose to take the granularity to the level of the pixels composing the image. In other words, each section can have as many subsections as there are pixels in it. Thus,  for any image composed of pixels color symmetries can be designed, for instance  to pre-existing source images  such as paintings. 
 
\noindent
  Let us consider images composed of pixels where each pixel $p_k$ has a color $c_k$ in a HSV color space: $c_k=(h_k,s_k,v_k)$. We start with focusing on  the hue $h_k$, assuming that the saturation $s_k$ and the value (brightness) $v_k$ remain constant for now. The method is equally applicable to the other elements of the HSV color space and in the visual results also such mappings are used. 
 We define a color permutation associated with the symmetry group by a map $f(h_k)$, which is a mapping of the hue component of the HSV color space onto itself. A similar mapping of the color space has recently been used to visualize number-theoretic functions with portraits of mathematicians~\cite{spec20}. Here, I use general algebraic functions solely to achieve certain color distortion effects. 
Thus, by using such a  map $f(h_k)$ and for rectangular source images, a color symmetry  
can be imposed by a color permutation as follows. For each subsection $\ell$, $\ell=1,2,\ldots,\Lambda$,  we set a map $f(h_k,\ell)$, which maps the hue component of the $k$-th pixel in subsection $\ell$, and thus changes the color of the pixel. If the color symmetry stipulates that the color is preserved, the map is the identity  
$f(h_k,\ell)=h_k$.
For instance, a color permutation associated with the rotation $\alpha$ can be expressed by a two-line notation as \begin{equation}
    p_\alpha=\left(\begin{smallmatrix} f_i(h_k,\ell_p)   \\ f_j(h_k,\ell_q)   \end{smallmatrix} \right) \quad \forall \{\ell_p,\ell_q\} \in \Lambda_\alpha, \quad \ell_p \neq \ell_q, \label{eq:col_per11}
\end{equation} where $\Lambda_\alpha$ is the set of rotation-symmetric subsections. 
A
rotation maps (\textbf{nw},\textbf{ne}) $\leftrightarrow$ (\textbf{se},\textbf{sw}). Thus, the color permutation \eqref{eq:col_per11}
 implies that the hue $h_k$ of all pixels $p_k$ in the $\ell_p$-th subsection in either \textbf{nw} or \textbf{ne} take the colors specified by $f_i(h_k)$ while all pixels $p_k$ in the rotation-symmetric $\ell_q$-th subsection in either \textbf{se} or \textbf{sw} take the colors specified by $f_j(h_k)$, and vice versa.  As this means that for any pair of colors $f_i(h_k)$ and $h_j(h_k)$, a rotation is taking the color $f_i(h_k)$ to the color $h_j(h_k)$, such a color symmetry is even a transitive coloring~\cite{roth82}.  If, as a simplification,  the upper map in \eqref{eq:col_per11} is the identity  
$f_i(h_k,\ell_p)=h_k$, then the color of the $\ell_p$-th subsection are identical to the source image, while the color of the symmetric subsection changes to $f_j(h_k,\ell_q)$. 
For the reflections, the color permutations can be defined likewise. Such a definition of a color permutation provides an algorithmic template for symmetry-consistent color distortions of source images.

\begin{figure}[h!tbp]
\centering
	\includegraphics[trim = 40mm 30mm 40mm 25mm,clip,width=4cm]{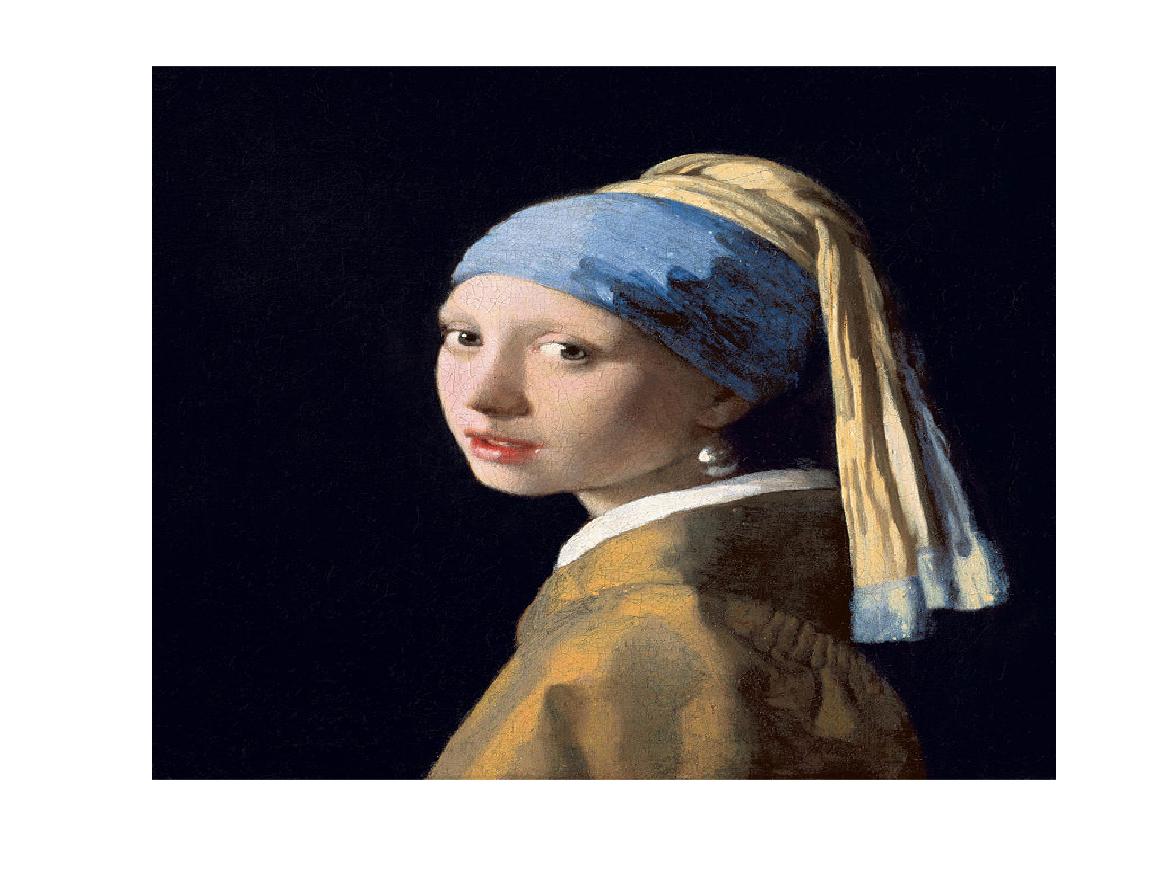}
    	\includegraphics[trim = 40mm 30mm 40mm 25mm,clip,width=4cm]{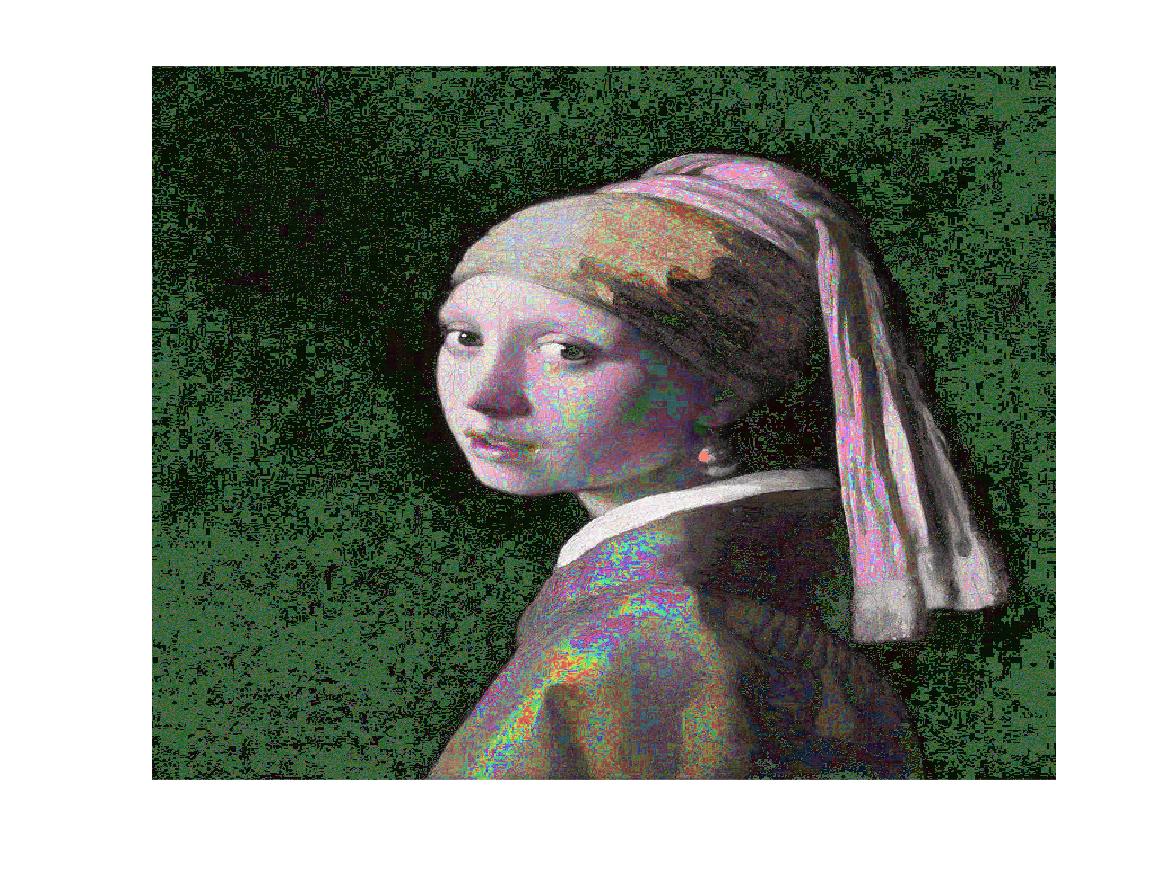}
	\includegraphics[trim = 40mm 30mm 40mm 25mm,clip,width=4cm]{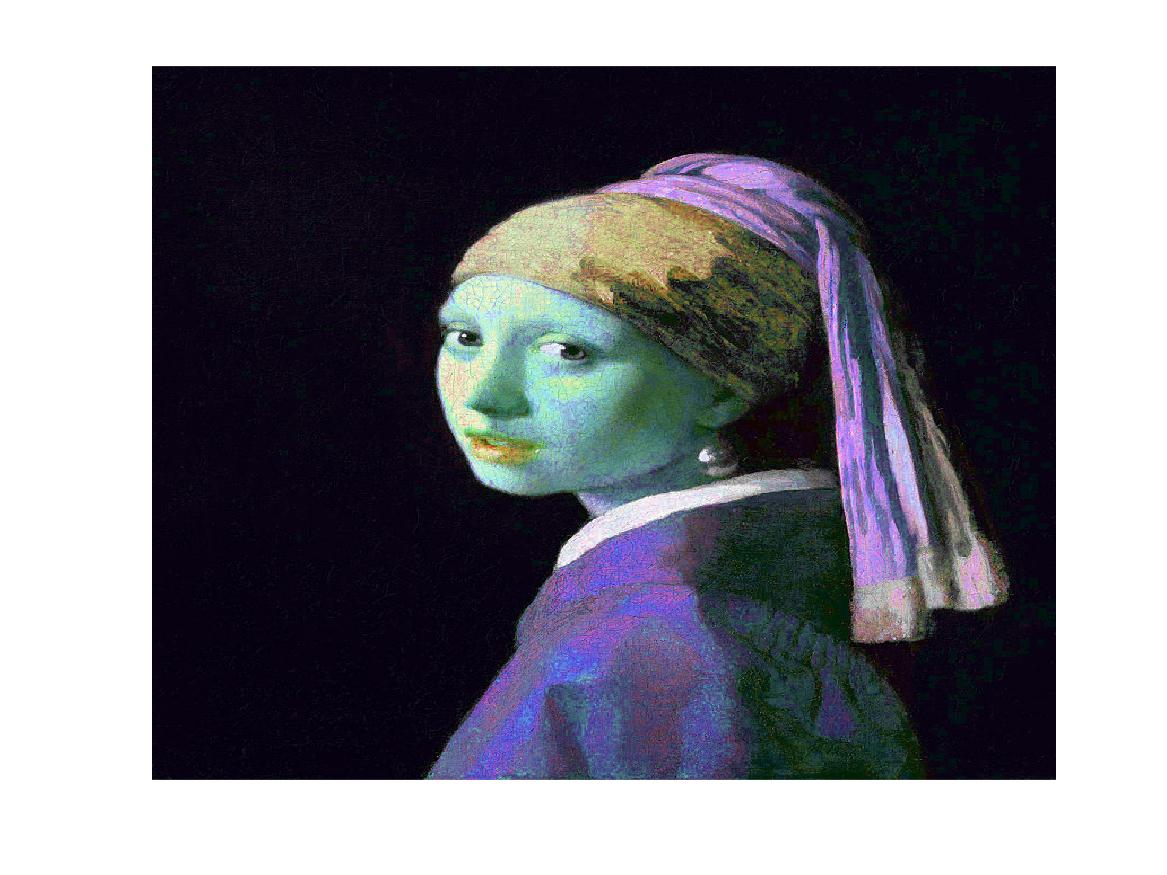}
		\includegraphics[trim = 40mm 30mm 40mm 25mm,clip,width=4cm]{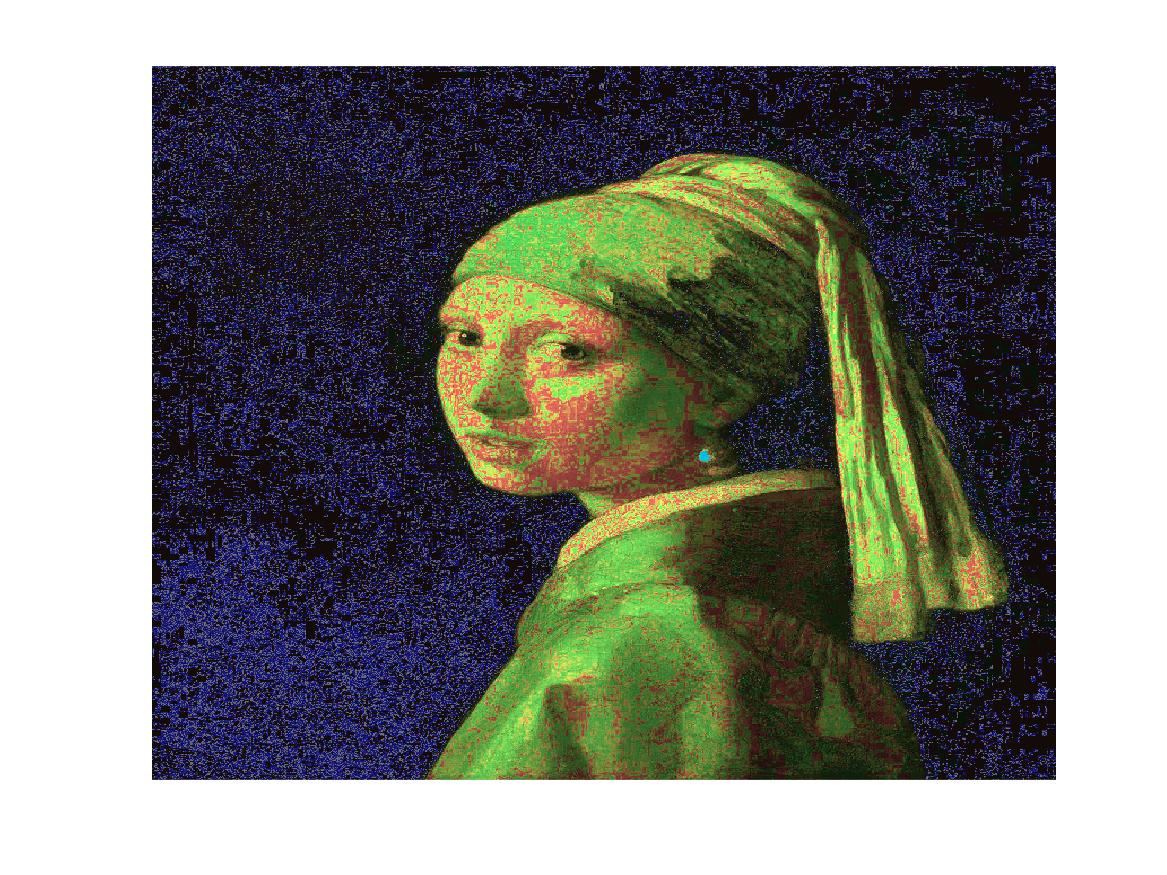}
		
			\includegraphics[trim = 40mm 30mm 40mm 25mm,clip,width=4cm]{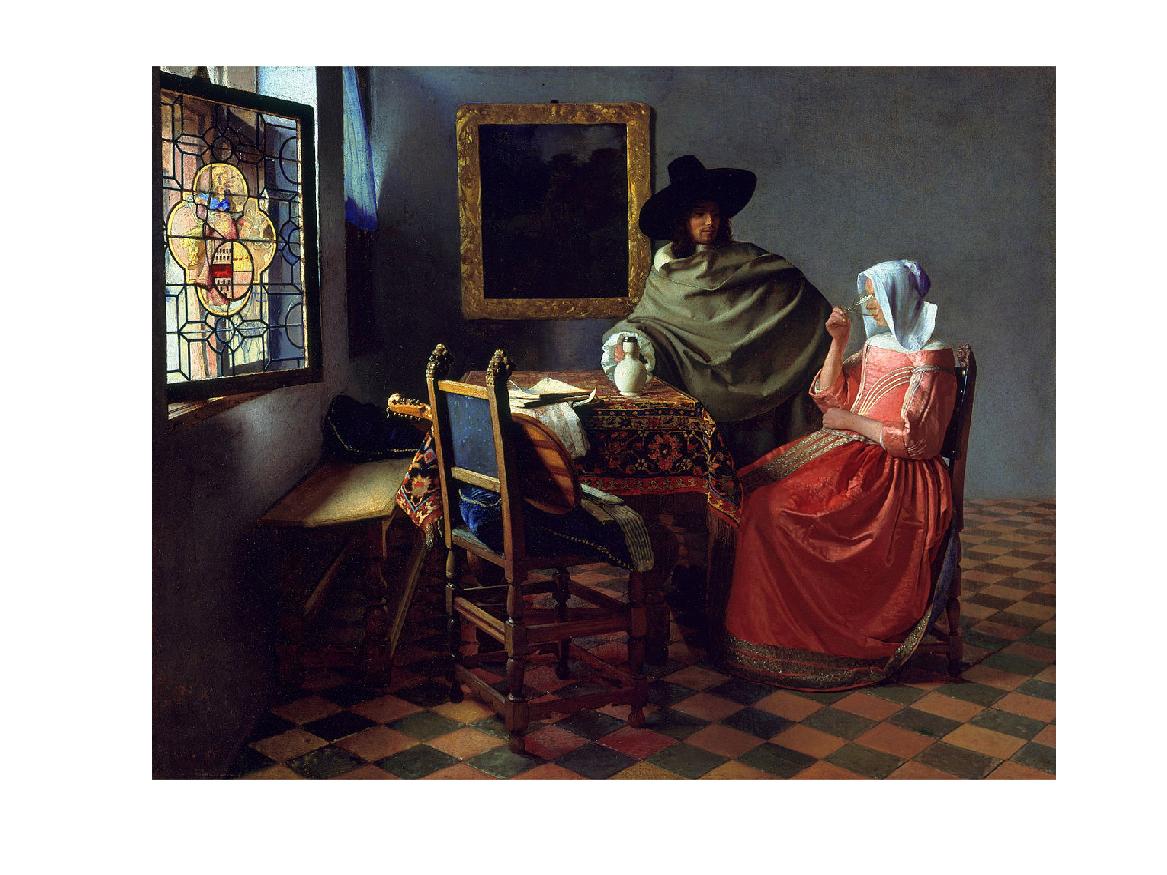}
    	\includegraphics[trim = 40mm 30mm 40mm 25mm,clip,width=4cm]{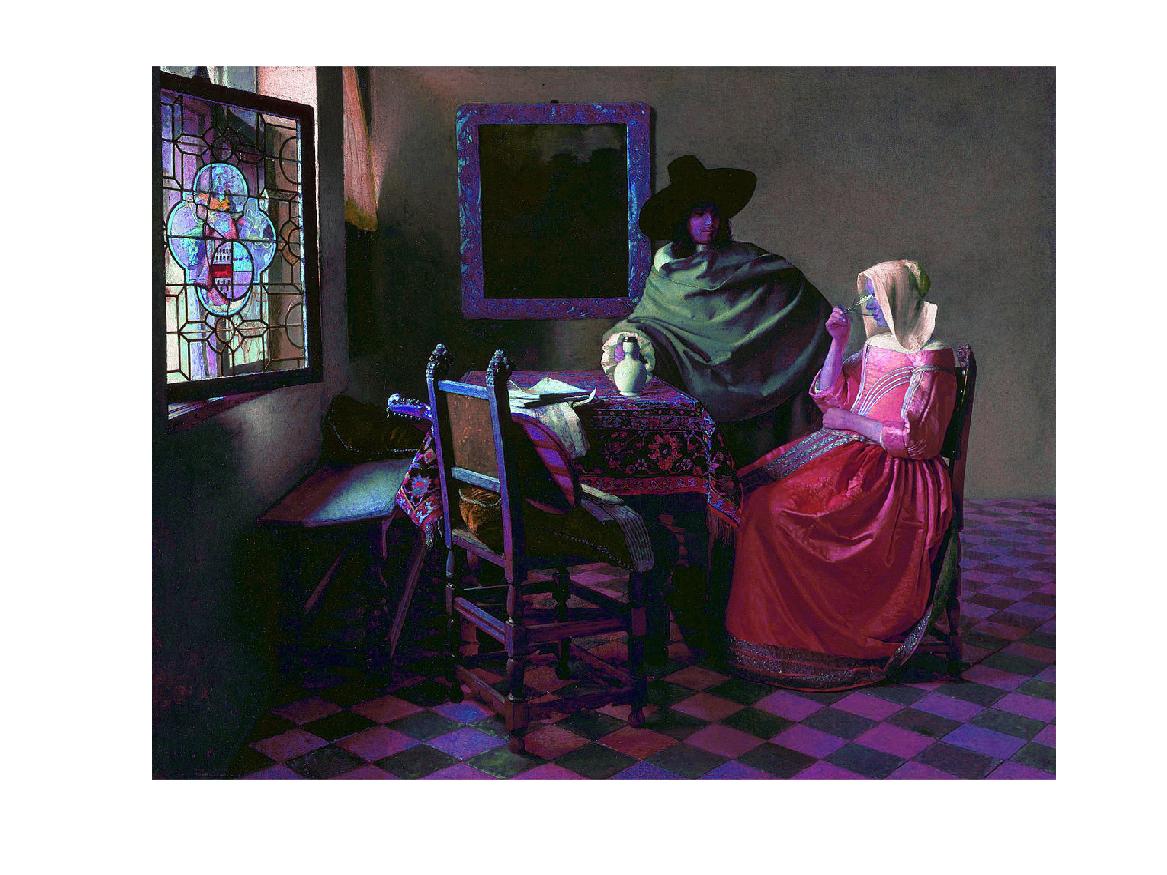}
	\includegraphics[trim = 40mm 30mm 40mm 25mm,clip,width=4cm]{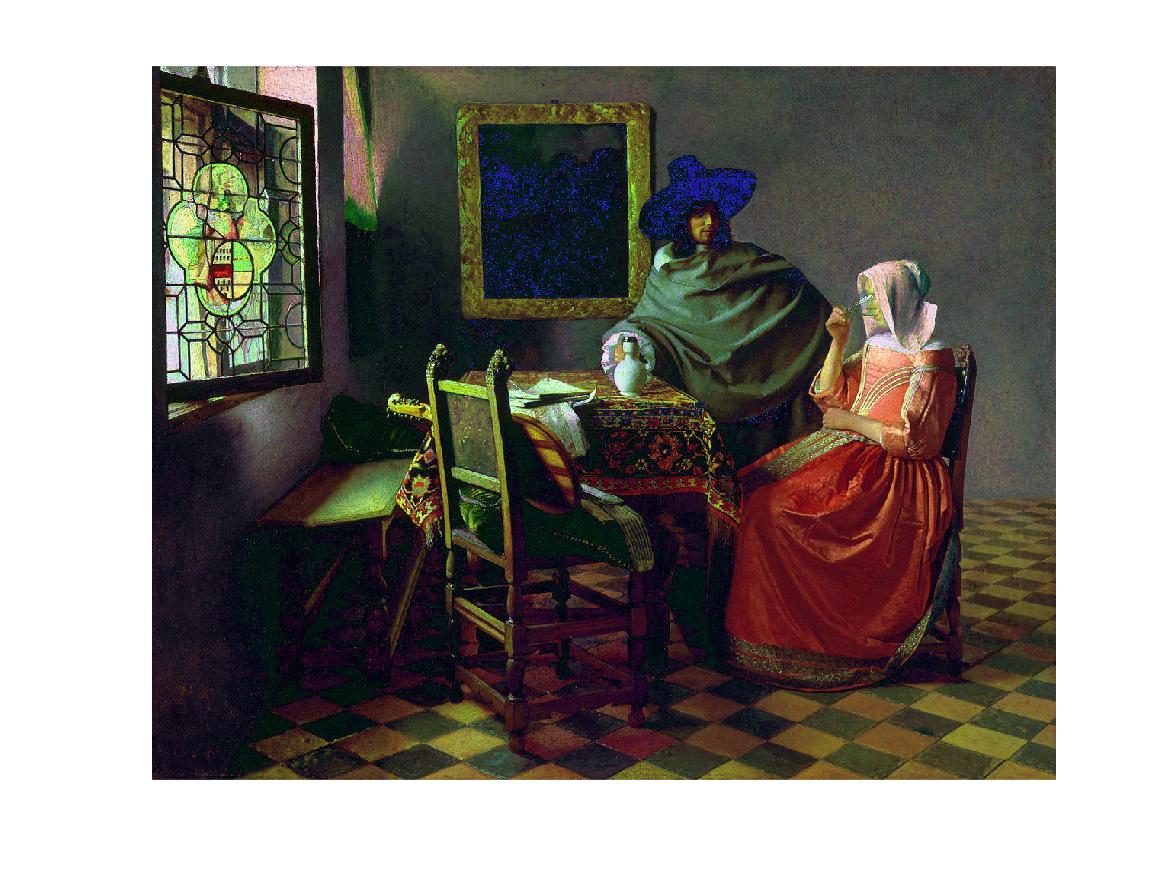}
		\includegraphics[trim = 40mm 30mm 40mm 25mm,clip,width=4cm]{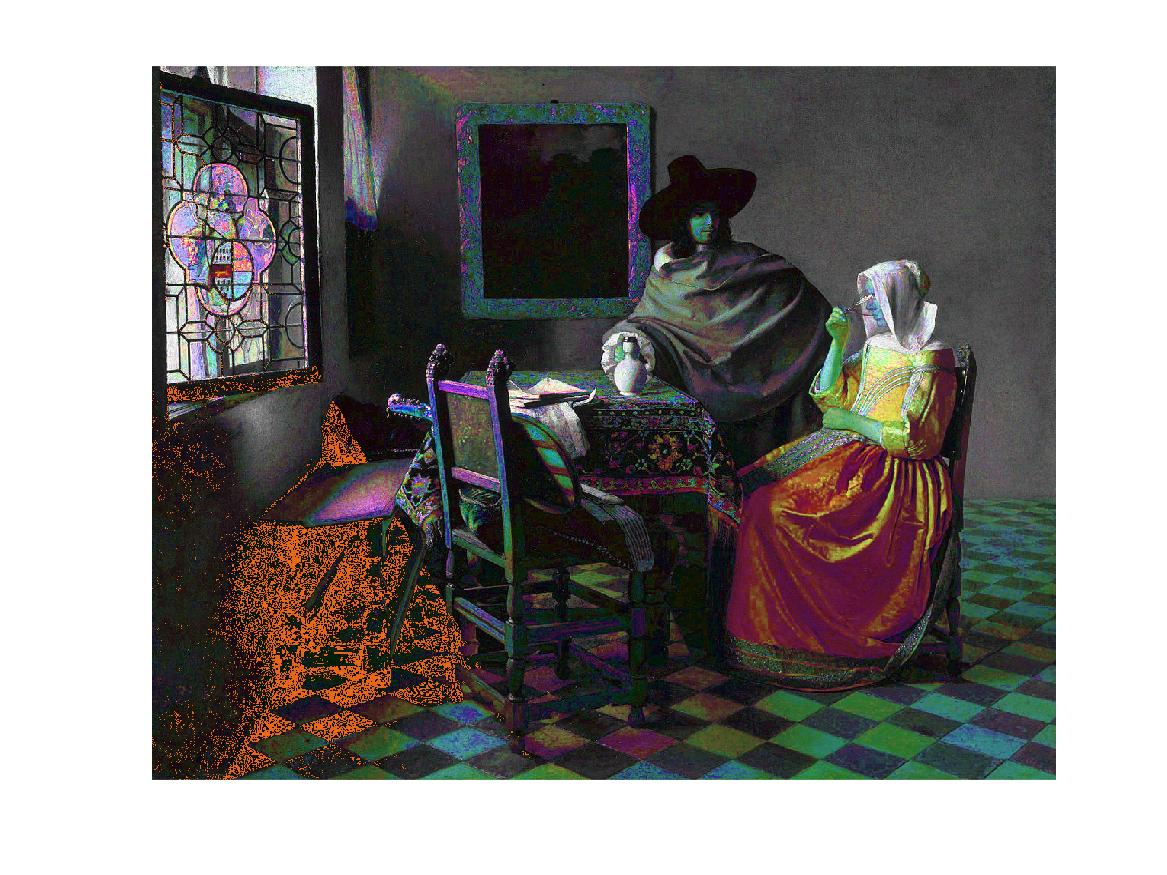}
		
				\includegraphics[trim = 40mm 30mm 40mm 25mm,clip,width=4cm]{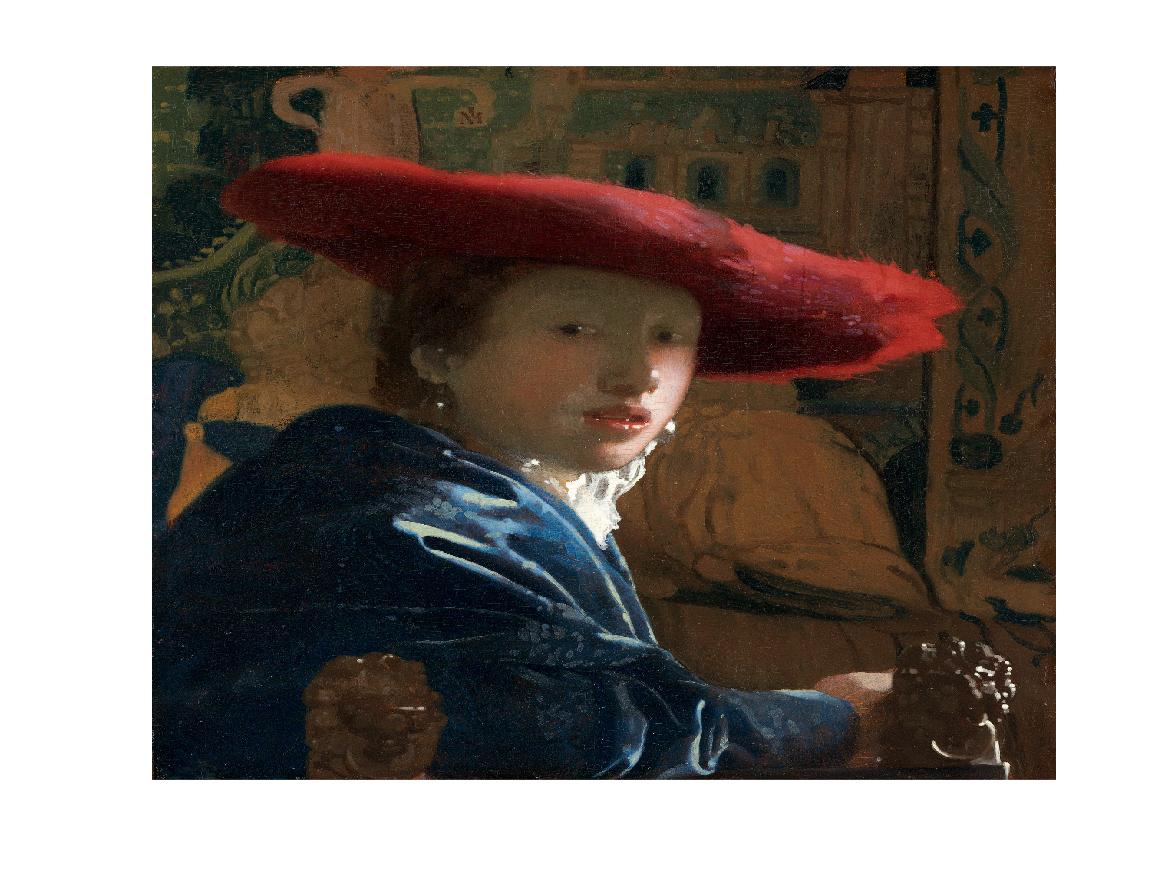}
    	\includegraphics[trim = 40mm 30mm 40mm 25mm,clip,width=4cm]{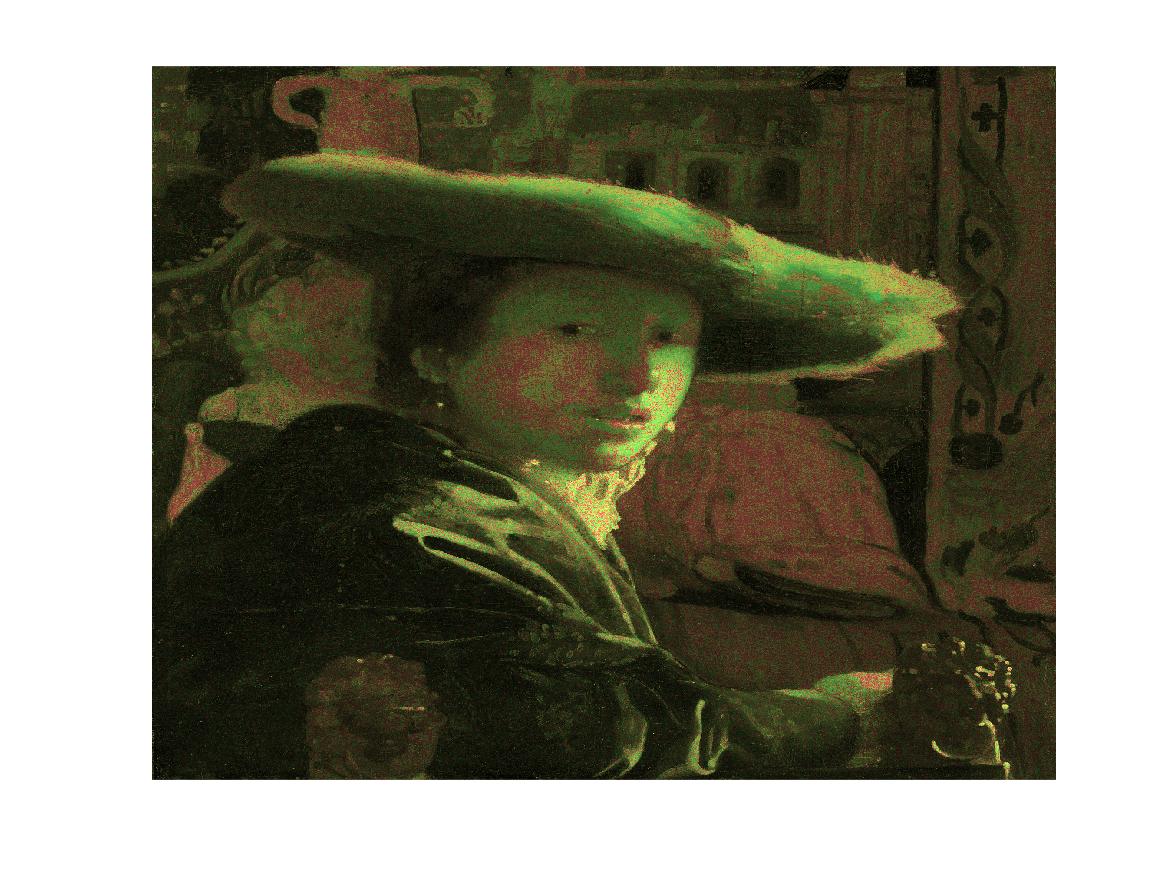}
	\includegraphics[trim = 40mm 30mm 40mm 25mm,clip,width=4cm]{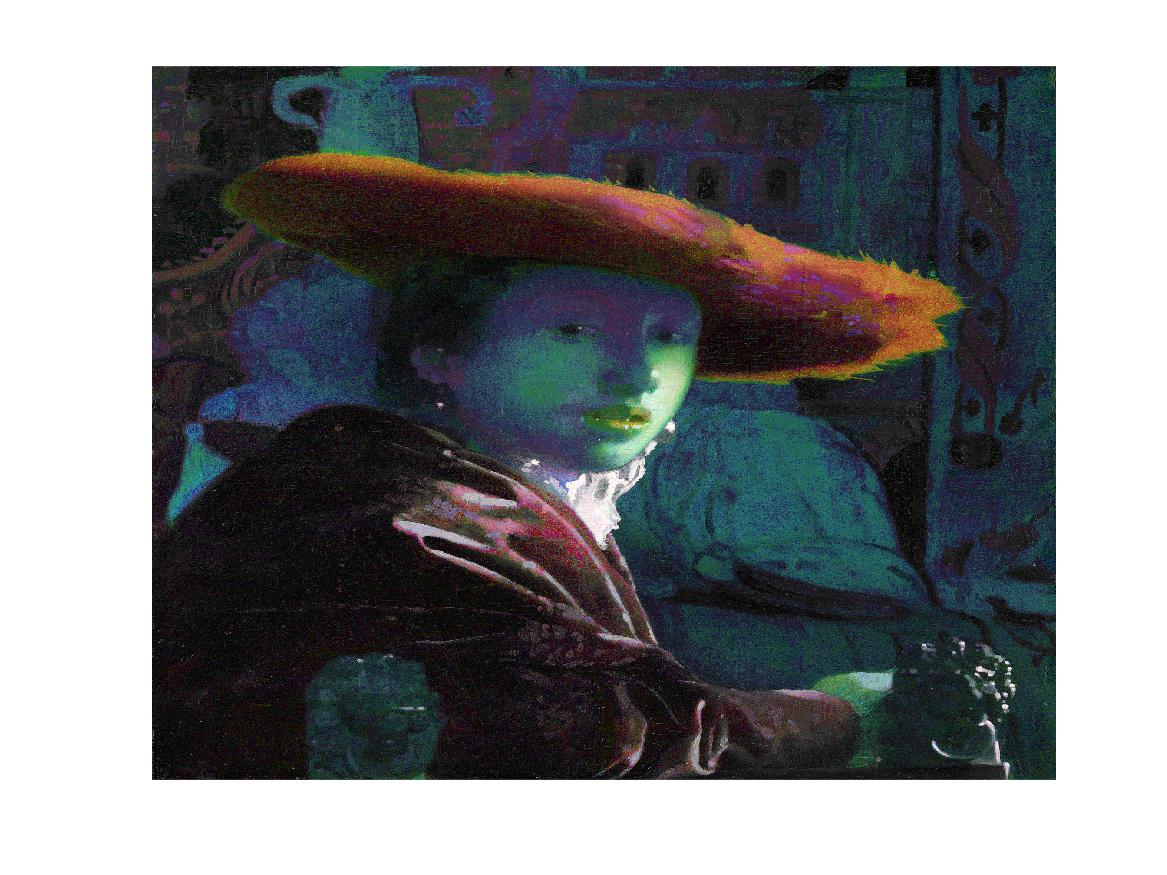}
		\includegraphics[trim = 40mm 30mm 40mm 25mm,clip,width=4cm]{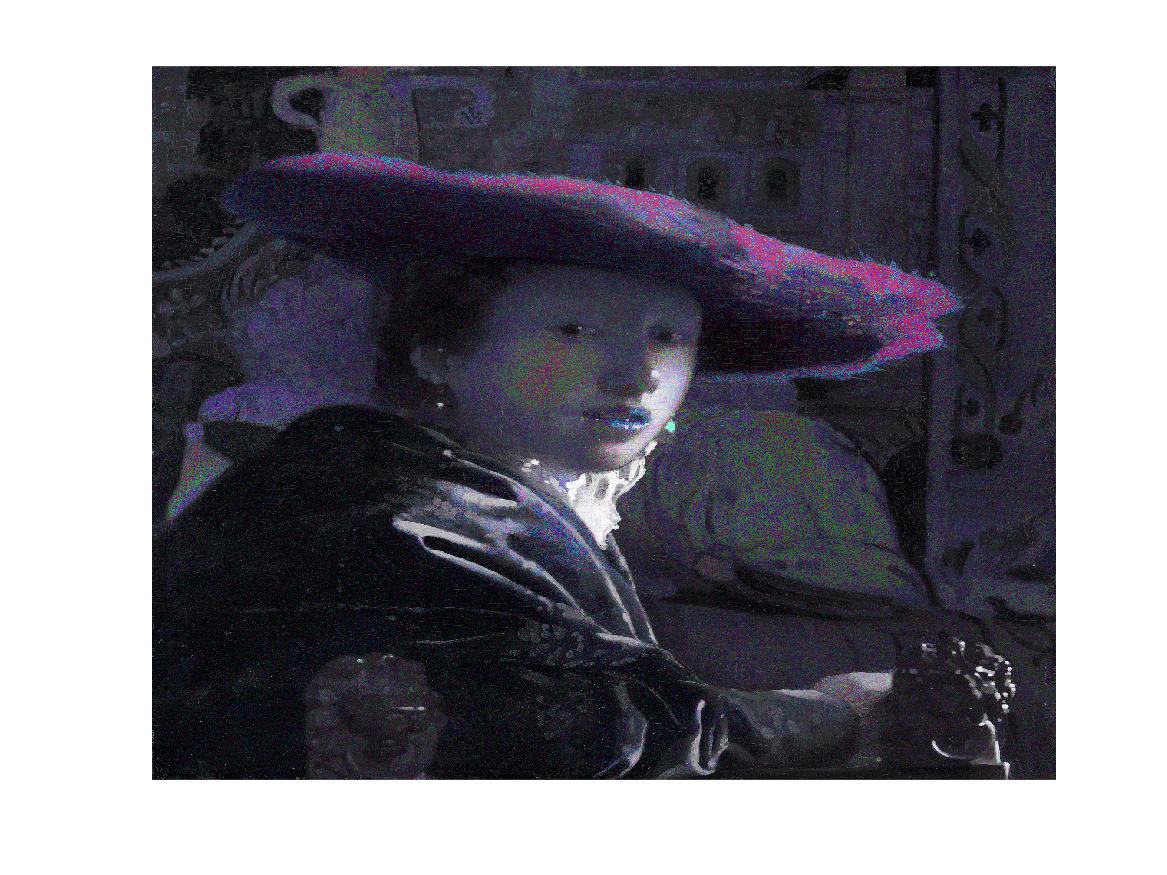}
		
\caption{Vermeer's {\it Girl with a Pearl Earring},
1665 (upper left) and {\it The Glass of Wine},
1661 (middle left) and {\it Girl with a Red Hat}, 1666 (lower left) with three  color distortions each. }
\label{fig:girl}
\end{figure}

\section*{Visual Results}

In this section, the color permutations associated with the rectangular symmetry group are applied and visual results are presented. As source images three paintings by Johannes Vermeer (1632-1675) are taken. Apart from
reasons of artistic taste
these paintings are particularly suitable for the color symmetric image distortions I intended to study.   Many Vermeer paintings feature fine-scaled and strong  contrasts in color, as for instance analytically shown for the {\it Girl with a Pearl Earring}~\cite{del20,van19}.
The maps varying the hue consequently create
a substantial variety of color shifts in the sections affected by color symmetry, particularly for the color maps involving trigonometric functions. Thus, fine-scaled and strong  contrasts in color lead to substantial color distortions. 
Moreover, 
if at least the pixels' saturation or brightness are preserved, the ``color symmetric'' sections of the image resembles the original image, but we also have a kind of alienation or distancing effect in terms of color. Results of the color changing distortions are shown in Figure~\ref{fig:girl}.
 As color maps the functions $f_1(h_k)=0.45 |\sin{(\sqrt{2}\cdot 20 \pi h_k})|+0.55 |\sin{(20 \pi h_k)}|$, $f_2(h_k)=0.5(1+\sin{(40 \pi h_k)})$ and $h_3(h_k)=4h_k(1-h_k)$, 
  $f_4(h_k)=4h_k(h_k-1)+1$, $f_5(h_k)=0.15(1+\cos{(40\pi h_k)})+0.5h_k$ and $f_n(h_k)=n h_k \mod 1$ are used. In addition, there are some instances where similar functions are employed to map value or saturation, for instance to change the dark background in the upper images of Figure~\ref{fig:girl}. 
The images obtained still resemble the Vermeer paintings, but also leave behind the impression of somehow wrong colors.
\begin{figure}[htb]
\centering
	\includegraphics[trim = 40mm 30mm 40mm 25mm,clip,width=5.25cm]{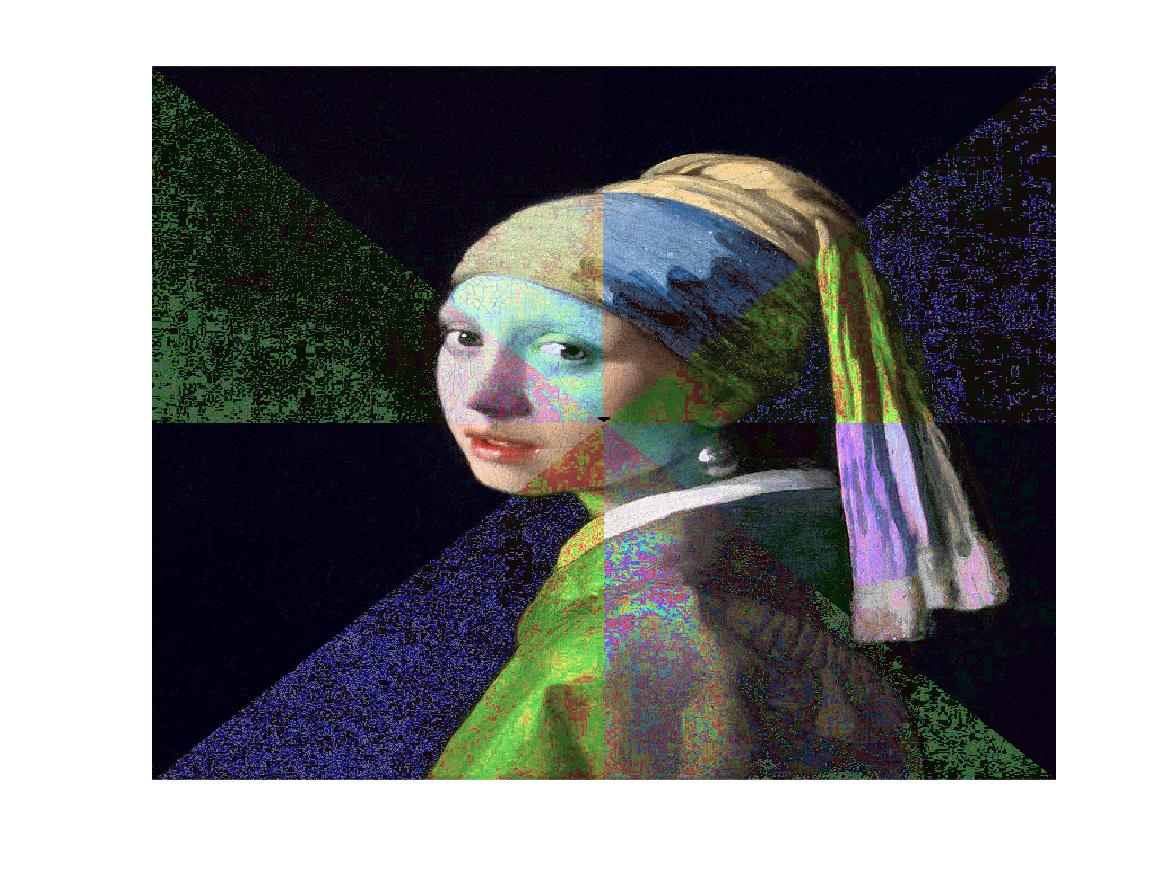}
    	\includegraphics[trim = 40mm 30mm 40mm 25mm,clip,width=5.25cm]{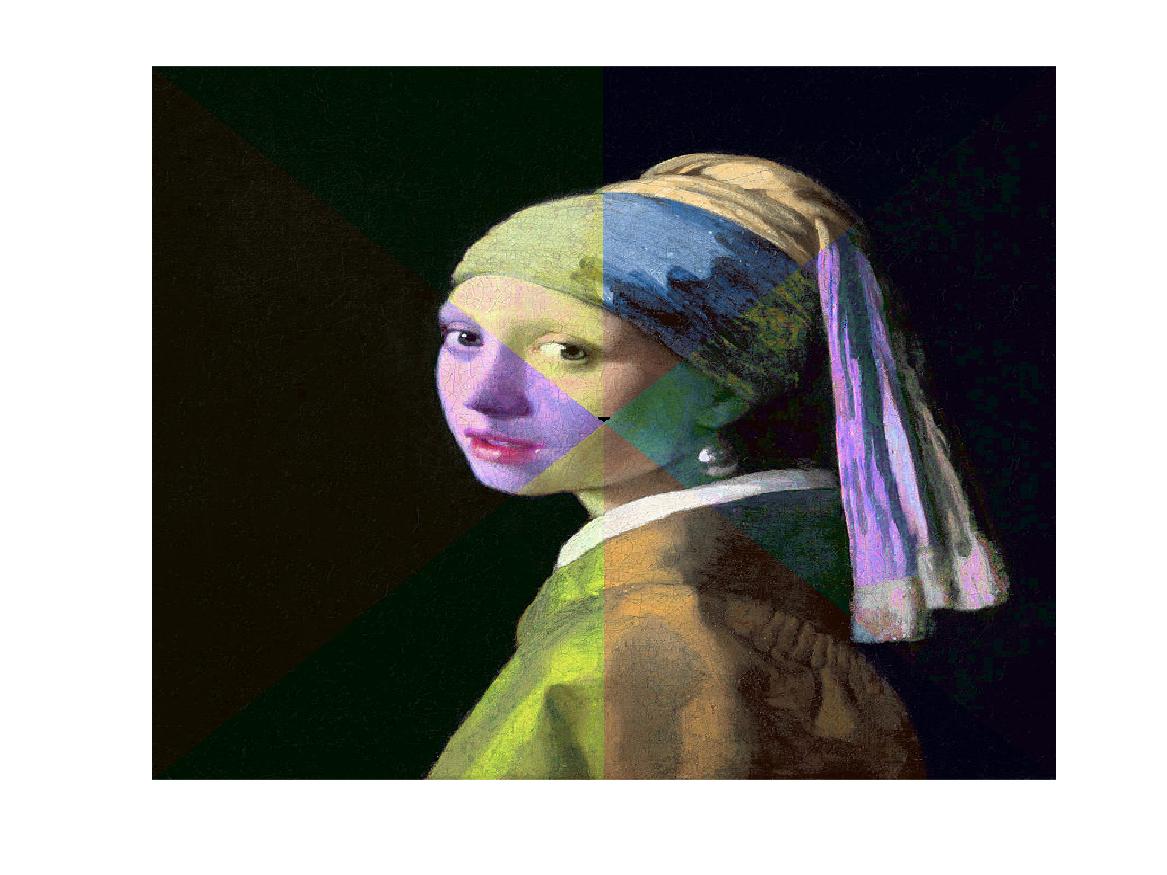}
	\includegraphics[trim = 40mm 30mm 40mm 25mm,clip,width=5.25cm]{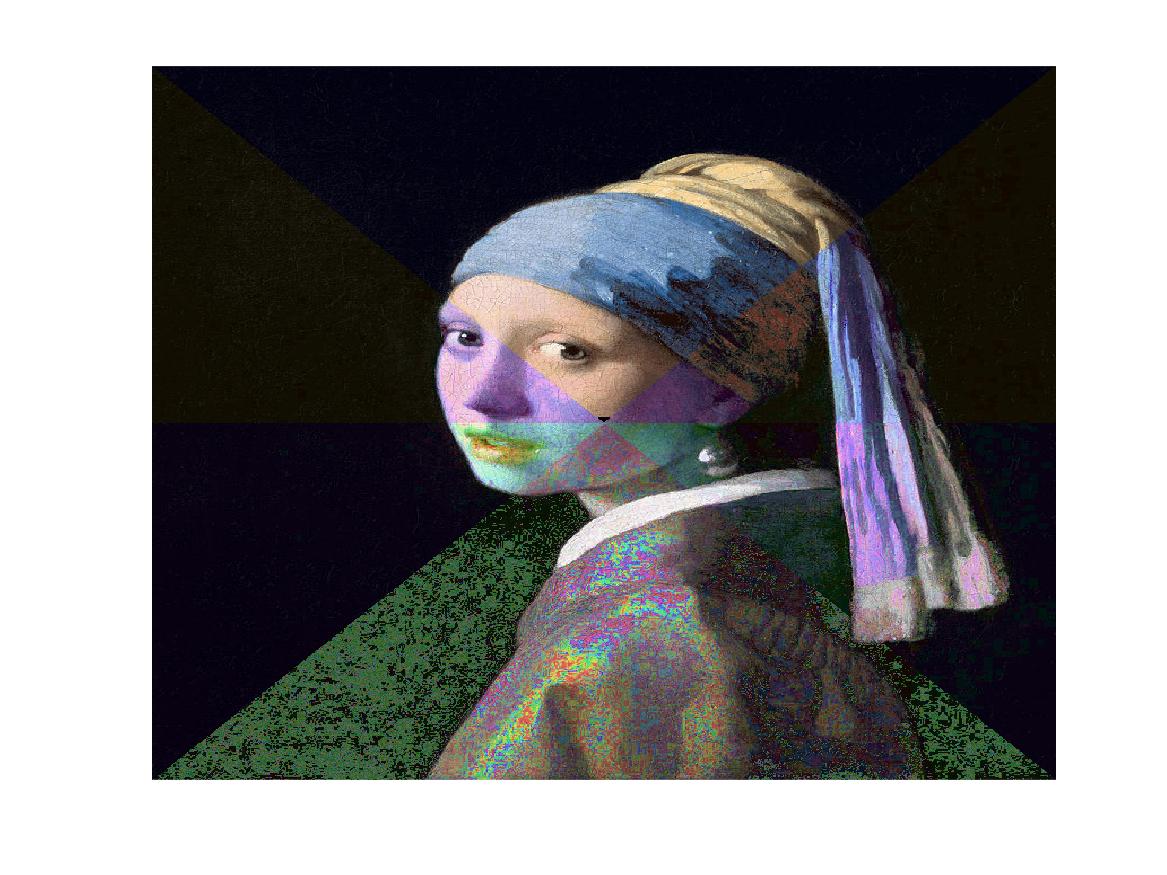}
	
		\includegraphics[trim = 40mm 30mm 40mm 25mm,clip,width=5.25cm]{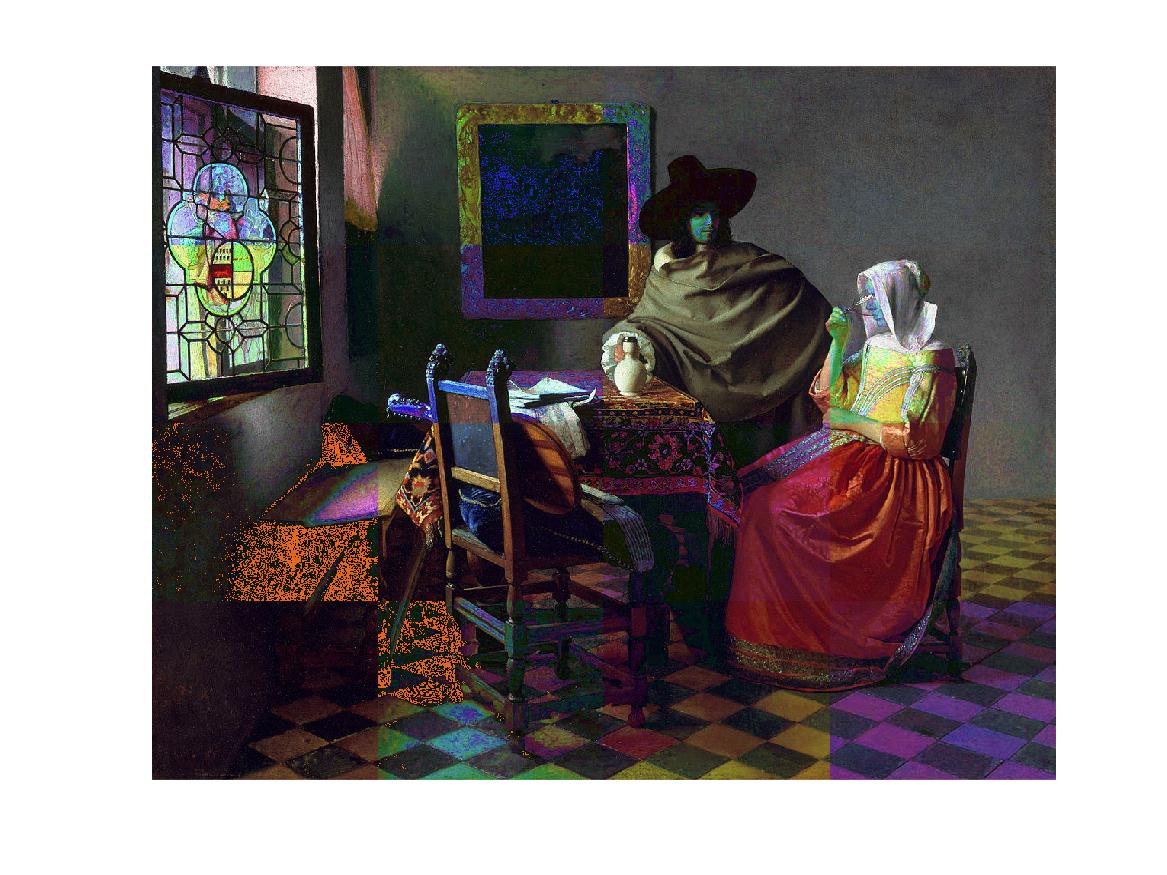}
    	\includegraphics[trim = 40mm 30mm 40mm 25mm,clip,width=5.25cm]{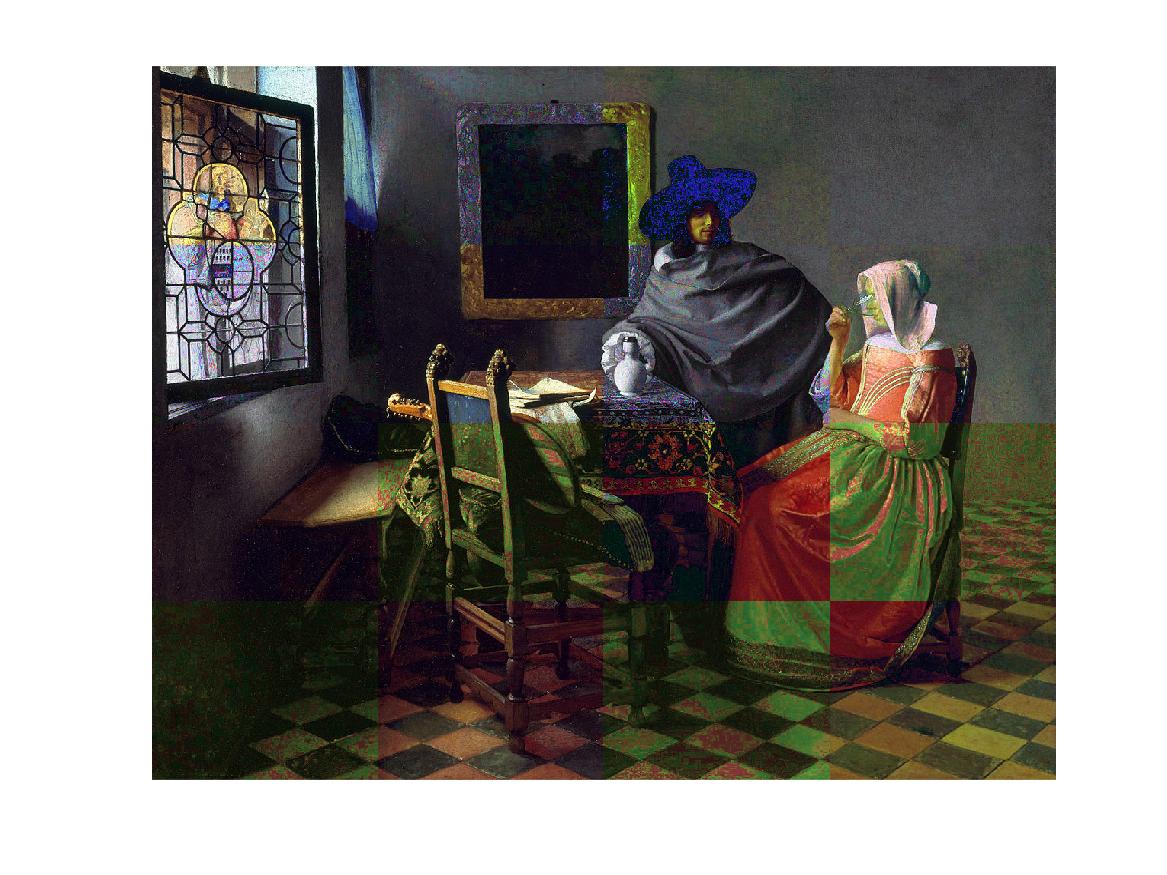}
	\includegraphics[trim = 40mm 30mm 40mm 25mm,clip,width=5.25cm]{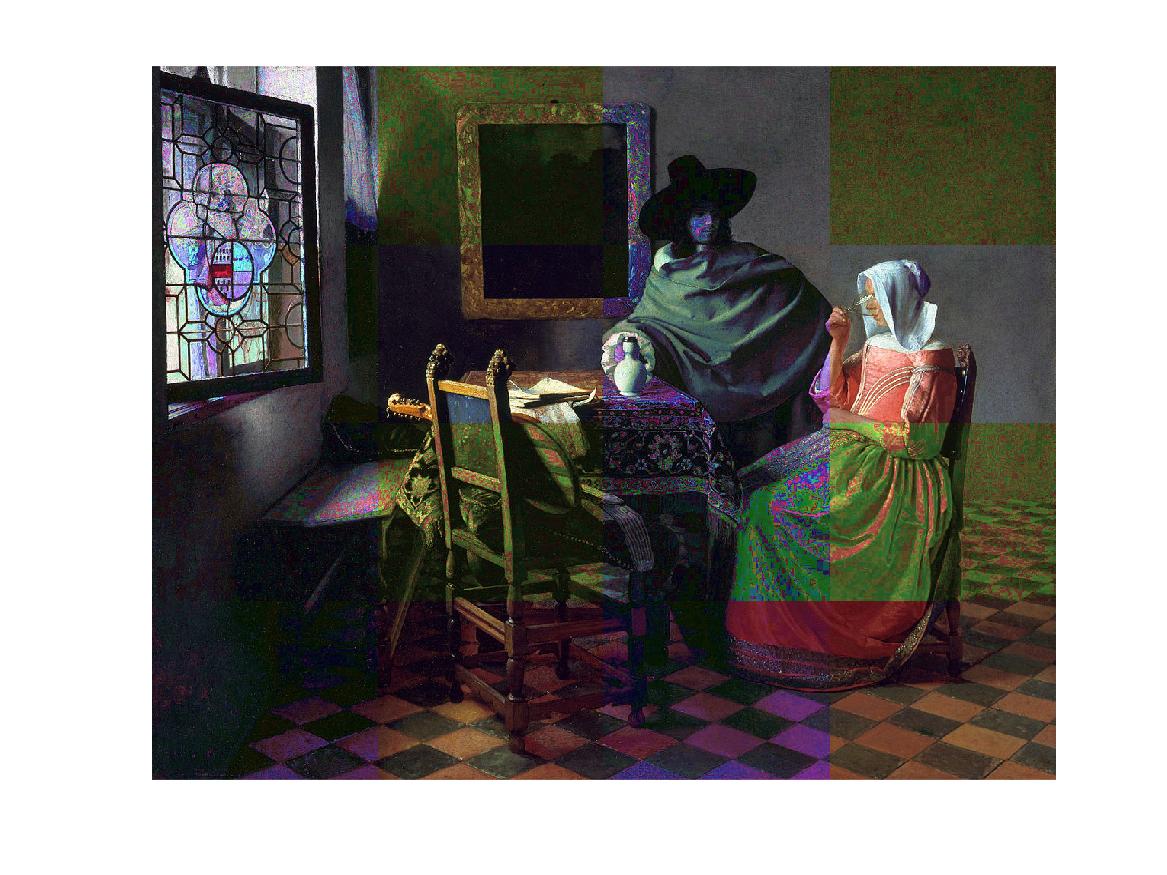}
	
		\includegraphics[trim = 40mm 30mm 40mm 25mm,clip,width=5.25cm]{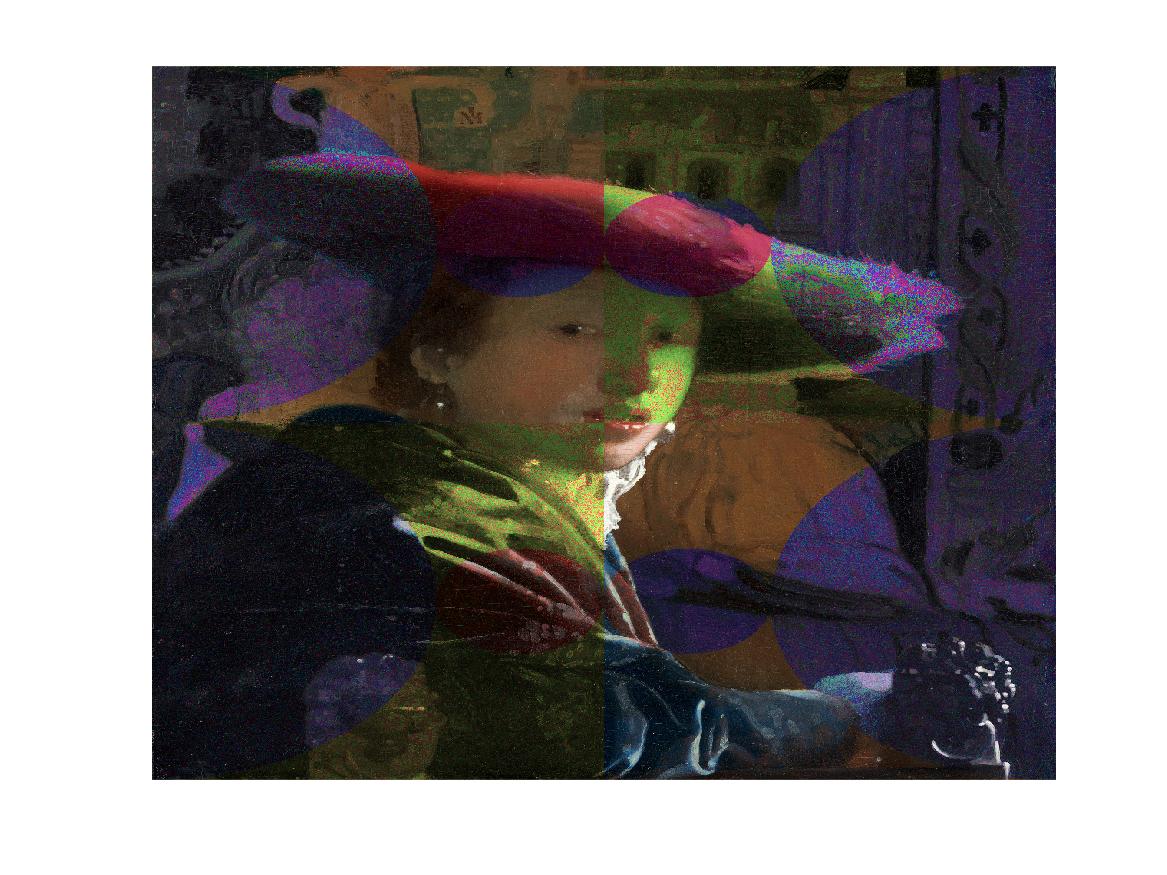}
    	\includegraphics[trim = 40mm 30mm 40mm 25mm,clip,width=5.25cm]{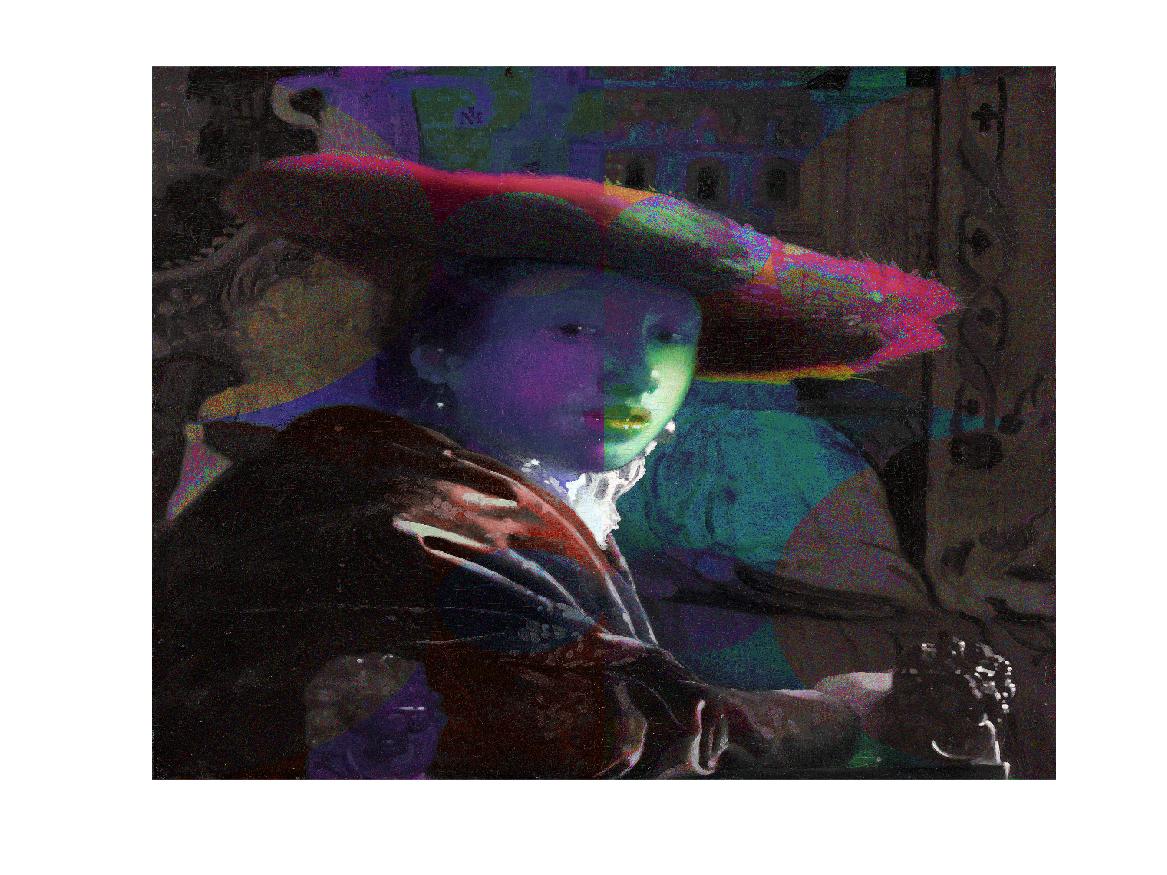}
	\includegraphics[trim = 40mm 30mm 40mm 25mm,clip,width=5.25cm]{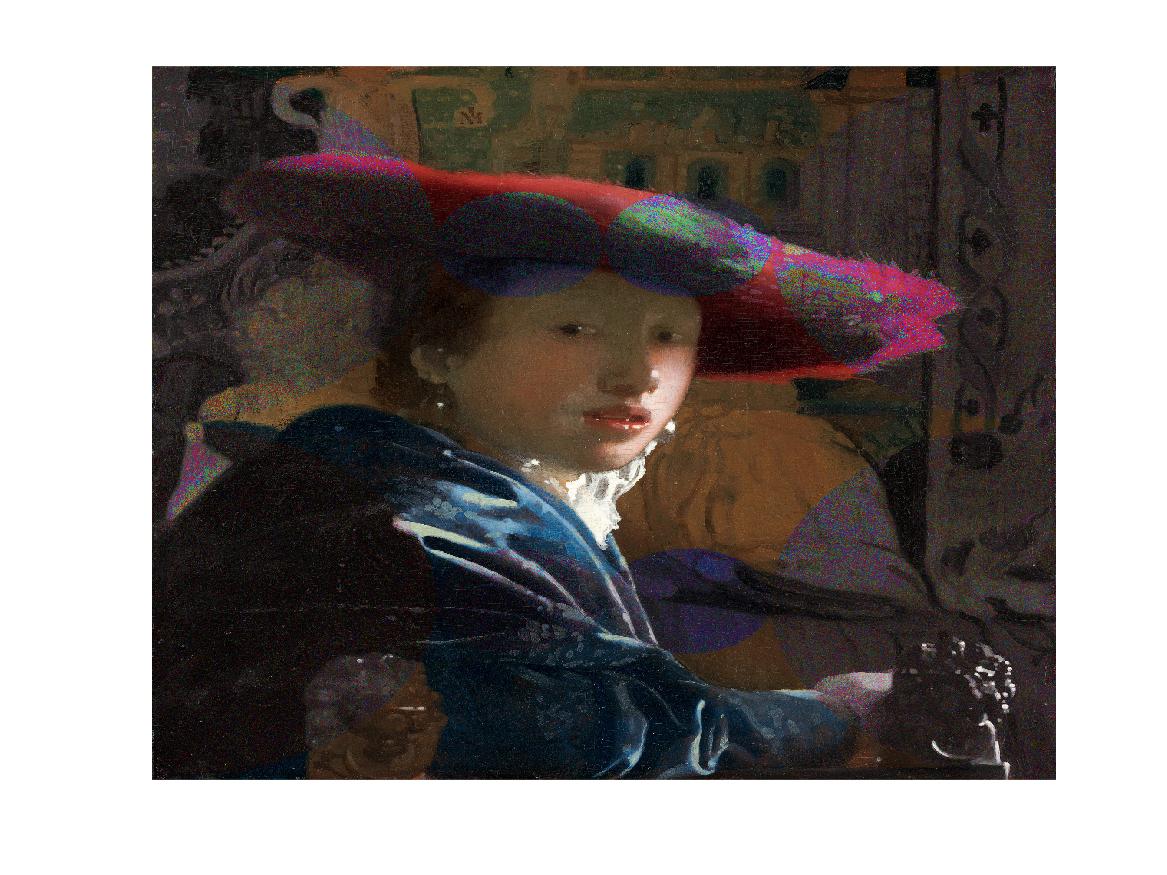}
	
\caption{Wrong colored Vermeer: Color-symmetric image distortions using the images in Fig. \ref{fig:girl}.}
\label{fig:vermeer_sym}
\end{figure}
Figure~\ref{fig:vermeer_sym} presents color symmetric image distortions using the Vermeer paintings. In each image, either a rotation, or a horizontal or a vertical reflection is realized. Subsections as in Figure~\ref{fig:simplecolorsymmetry} are used to create symmetric color effects, namely triangular ({\it Girl with a Pearl Earring}), chessboard-line ({\it The Glass of Wine}) and bubble-like ({\it Girl with a Red Hat}) subsection. Thus, works of
 generative art emerge in which can be seen as quotations, homages and reinterpretations of
 Vermeer's famous works.

\end{document}